\documentclass[runningheads]{llncs}
\usepackage[T1]{fontenc}
\usepackage{graphicx}
\usepackage{hyperref}
\hypersetup{
    colorlinks=true,
    linkcolor=red,
    urlcolor=magenta,
    citecolor=blue
}

\usepackage{amsmath}
\usepackage{amssymb}
\usepackage{algpseudocode}
\usepackage{algorithm}
\usepackage{booktabs}
\usepackage{dblfloatfix}
\usepackage{cite}
\usepackage[dvipsnames]{xcolor}
\usepackage{soul}

\begin{document}

\title{Graph-based Point Cloud Surface Reconstruction using B-Splines
%\thanks{Supported by University of Antwerp.} 
}
%
%\titlerunning{Abbreviated paper title}
% If the paper title is too long for the running head, you can set
% an abbreviated paper title here
%
\author{Stuti Pathak \and
Rhys G. Evans\and
Gunther Steenackers \and
Rudi Penne}
\authorrunning{S. Pathak et al.}
% First names are abbreviated in the running head.
% If there are more than two authors, 'et al.' is used.
%
\institute{University of Antwerp, Belgium}
\maketitle              % typeset the header of the contribution
\begin{abstract}
Generating continuous surfaces from discrete point cloud data is a fundamental task in several 3D vision applications. Real-world point clouds are inherently noisy due to various technical and environmental factors. Existing data-driven surface reconstruction algorithms rely heavily on ground truth normals or compute approximate normals as an intermediate step. This dependency makes them extremely unreliable for noisy point cloud datasets, even if the availability of ground truth training data is ensured, which is not always the case. B-spline reconstruction techniques provide compact surface representations of point clouds and are especially known for their smoothening properties. However, the complexity of the surfaces approximated using B-splines is directly influenced by the number and location of the spline control points. Existing spline-based modeling methods predict the locations of a fixed number of control points for a given point cloud, which makes it very difficult to match the complexity of its underlying surface. In this work, we develop a Dictionary-Guided Graph Convolutional Network-based surface reconstruction strategy where we simultaneously predict both the location and the number of control points for noisy point cloud data to generate smooth surfaces without the use of any point normals. We compare our reconstruction method with several well-known as well as recent baselines by employing widely-used evaluation metrics, and demonstrate that our method outperforms all of them both qualitatively and quantitatively.

\end{abstract}

% \vspace{10cm}

\section{Introduction}

Advanced technologies such as LiDAR, photogrammetry and structured light scanning have made the direct acquisition of discrete representations of real-world objects and physical environments around us more accessible and efficient. These 3D representations, formally known as point clouds, are inherently unstructured, contain outliers and are quite noisy due to sensor limitations, multi-view scanning, occlusions and environmental factors. The generation of continuous smooth surface models from these point clouds is extremely crucial for applications in diverse fields like CAD \cite{azernikov2004efficient}, robotics \cite{mittendorfer20123d}, medical imaging \cite{xiong2022automatic}, cultural heritage preservation \cite{gomes20143d}, etc. Therefore, the challenging task of reconstructing accurate surfaces from raw point clouds has become an increasingly active area of research in recent years \cite{huang2024surface, berger2017survey, farshian2023deep}.

\begin{figure}[H]
\includegraphics[width=\textwidth, trim=0cm 66cm 43cm 0cm, clip]{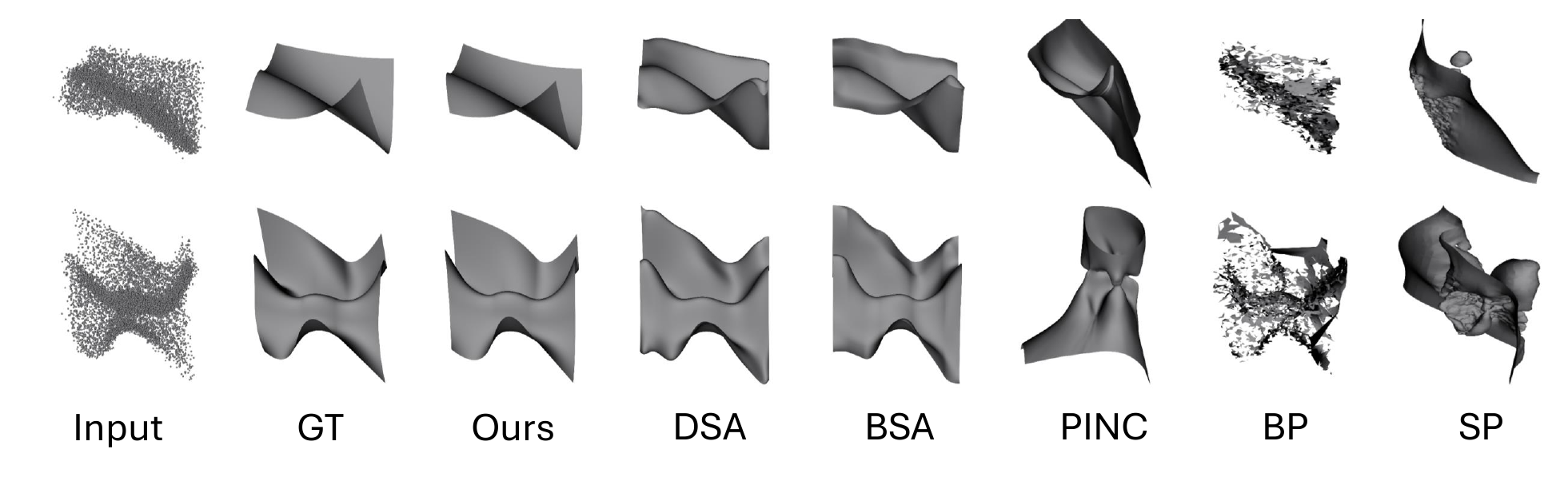}
\vspace{-0.3cm}
           \caption{Surface reconstruction of two noisy point clouds using our method and several baselines: Ball-Pivoting (BP) \cite{bernardini2002ball}, Screened Poisson (SP) \cite{kazhdan2013screened},  Least Square B-spline Surface Approximation (BSA) \cite{pieql1997nurbs}, Differentiable Spline Approximations (DSA) \cite{cho2021differentiable}, and p-Poisson Surface Reconstruction in Curl-Free Flow (PINC) \cite{park2023p}. This demonstrates our method's superior visual quality when compared to other approaches.}
  \label{fig:first}

\end{figure}

% \cite{cho2021differentiable, ben2020deepfit} are dsa and deepfit respectively
% pieql1997nurbs is geomdl

Surface reconstruction has evolved significantly over the decades, transitioning from classical geometric techniques such as Delaunay triangulation-based methods \cite{amenta1998surface} and Marching cubes \cite{lorensen1998marching}, to more advanced approaches such as Poisson surface reconstruction \cite{kazhdan2006poisson} and complete or partial machine learning-based techniques \cite{cho2021differentiable, ben2020deepfit}. Spline-based surface approximation techniques \cite{pieql1997nurbs, yuwen2006b, wang2023reconstruction} stand out notably for their capacity to deliver surfaces that are inherently smooth, continuous and accurately represent the underlying geometry of a point cloud. Additionally, because spline surfaces are defined parametrically, they provide a more compact surface representation compared to dense mesh-based models.

% \cite{lenssen2020deep, li2022graphfit} are iternet(Deep Iterative Surface Normal Estimation) and graphfit respectively

Graph Convolutional Networks (GCNs) have emerged as a powerful paradigm in the field of learning-based computer vision, particularly in the context of point cloud data processing. Leveraging graph-based representations, GCNs enable the extraction of meaningful features, capture the spatial dependencies, and learn complex patterns within unstructured point clouds \cite{wang2019dynamic}. Over time, several works have explored their relevance for 3D vision tasks such as object recognition \cite{landrieu2018large}, segmentation \cite{shi2020point}, etc. In the context of surface reconstruction, a few works have employed GCNs but have reformulated this task as one of normal estimation \cite{lenssen2020deep, li2022graphfit}. In these approaches, the surface geometry is subsequently derived from the predicted normals.

Regardless of the algorithm employed, surface reconstruction methods often face several persistent challenges, including generation of non-watertight meshes, erroneous and overlapping triangulations, self-occlusions, lack of consistent intermediate normal estimates, high sensitivity to noise as well as to algorithm-specific hyperparameters \cite{huang2024surface}. While traditional analytical methods often struggle to produce reliable meshes from noisy point clouds, machine learning-based approaches attempt to address this limitation by training on datasets with known ground truth normals or meshes \cite{ben2020deepfit, lenssen2020deep, li2022graphfit}. However, in real-world scenarios, point clouds are typically acquired without access to any ground-truth information. As a result, researchers commonly resort to using synthetic datasets with known ground truth values, or analytically estimated normals as surrogate ground truth, an approach that is itself highly unreliable when dealing with noisy point clouds. In both of the mentioned scenarios, the method generalizability and accuracy for out-of-distribution data reduces significantly. Therefore, there is a need for methods that can directly reconstruct smooth and accurate surfaces from noisy point clouds, without relying on ground-truth normals or their intermediate estimation, while effectively addressing the aforementioned challenges.

% \cite{pieql1997nurbs, cho2021differentiable} add more mlbased that fix spline size
% \cite{ben2020deepfit, lenssen2020deep, li2022graphfit} normal gt needed

A B-spline surface can be completely characterized by its degrees in the two parametric directions, and a grid of control points, whereas other elements such as knot vectors and basis functions can be systematically derived from these primary components. Several spline-based algorithms address the reconstruction task by predicting the location of a fixed number of control points for any given surface  \cite{pieql1997nurbs, cho2021differentiable, sharma2020parsenet}. However, the assumption of a fixed number of control points leads to underfitting when this number is insufficient, or overfitting when it exceeds what is necessary to capture the true complexity of the input point cloud (as illustrated in Figure \ref{fig:spline}). Furthermore, most of these methods work strictly on ordered grid-structured point sets, which makes them unsuitable for unordered point cloud datasets. Hence, a spline-based surface reconstruction method should be able to predict not only the locations of control points but also their appropriate number needed to represent the underlying surface of an unordered point cloud accurately.

% Moreover, learning-based articles \textbf{(cite some)} experiment only with simple surfaces, those for which control points overlap with the surface itself. 

% There have been works where people use other parametric surfaces (DeepFit graphfit) to fit point cloud patches but fix the degree of their n jet. This is an issue and it limits the

% -noisy
% -variabl cp
% -no normal

In this paper, we address the aforementioned challenges by implementing a graph-based dictionary-guided surface reconstruction technique which fits B-spline surfaces with variable control points to unordered noisy point clouds, without the need of any point normals. Our approach leverages a graph neural network to predict both the locations and the optimal number of control points for a given point cloud. Moreover, we create our own noise-induced point cloud dataset using B-spline modeling. We experiment on both our dataset as well as randomly generated surfaces. We compare our method against several recent and widely-used baselines, both qualitatively (as illustrated in Figure \ref{fig:first}) and quantitatively using standard evaluation metrics.

% The novelty of this paper lies in genarlizability to surfaces with intrinsically different number of control points which helps in modelling both complex and simple surfaces. 

% \textcolor{red}{The remaining paper is organised as follows. In Section \ref{related_work}, we discuss both traditional and }

\section{Related Work}
\label{related_work}

% \textcolor{red}{A wide range of methods have been developed for surface reconstruction from point clouds, as comprehensively reviewed in \cite{huang2024surface} and \cite{berger2017survey}.} 

Among the most widely adopted conventional surface reconstruction techniques are the Ball-Pivoting \cite{bernardini2002ball} and Screened Poisson \cite{kazhdan2013screened}. The former simulates a ball rolling over a point cloud, forming triangles whenever it touches three points simultaneously, thereby generating a mesh over the surface. The latter builds on the classic Poisson surface reconstruction \cite{kazhdan2006poisson} by introducing a dualized screening term to improve its geometric fidelity by explicitly incorporating the points of a point cloud as interpolation constraints. Due to their popularity, these methods have been integrated into various point cloud processing tools such as MeshLab \cite{LocalChapterEvents:ItalChap:ItalianChapConf2008:129-136}. Although providing accurate surfaces in case of non-noisy and uniformly distributed point clouds, these methods fail miserably when noise comes into play.

Spline-based surface reconstruction methods aim to generate smooth and continuous surfaces from point cloud data by leveraging on B-spline or NURBS (Non-Uniform Rational B-Splines) basis functions. Given a fixed number of control points, least square surface approximation technique using B-splines \cite{pieql1997nurbs} basically fits spline curves along the two directions of a point cloud. This method adjusts the control points mathematically to best match the spline surface with the input cloud while keeping the surface smooth. Recent papers have shown the applications of spline-based approaches in diverse fields such as industrial manufacturing \cite{wang2023reconstruction} and areal data analysis \cite{harmening2017choosing}. However, inherently, these algorithms not only require the manual tuning of the number control points but also demand a point cloud to be in an ordered grid structure.

With the advent of the machine learning era and the availability of huge point cloud datasets, data-driven surface reconstruction algorithms have emerged as powerful alternatives to traditional methods \cite{farshian2023deep}. A recent paper \cite{cho2021differentiable} extends the use of splines, for approximating functions, in differentiable machine learning models by deriving a weak Jacobian for these functions. They demonstrate surface reconstruction from point clouds as an application of their method. \cite{park2023p} proposes a p-Poisson equation-based reconstruction method which learns an approximated signed distance function with the given raw point cloud data. Unlike other machine learning approaches \cite{ben2020deepfit, lenssen2020deep, li2022graphfit}, these methods operate solely on the $xyz$-coordinates of point clouds, making them suitable baselines for comparison in this study.

All of the methods discussed above are either extremely sensitive to noise or the tuning of their algorithm's hyperparameters, such as the number of control points for spline-based techniques. In this work, we try to solve the problem of surface reconstruction by modeling the underlying geometry of a noisy point cloud using learnable graph connections and accurate spline fitting. By combining data-driven and analytical techniques, this approach offers an improvement over existing methods that rely solely on one or the other.

% \textbf{Doesn't need PC datasets with known normals. Cannot compare to a lot of recent ML models beacuse all of them use datasets with normals (cite all)}

\section{Background}
\label{sec:bg}
In this section, we present the mathematical definitions of B-spline curves and surfaces which will make the reader familiar with the idea B-spline modeling. Moreover, we highlight the shortcomings of B-spline fitting algorithms. 

B-spline, or basis spline, in its simplest form, is a parametric piece-wise polynomial function. A B-spline curve of degree $k$ with $n+1$ control points (CPs) $\mathbf{P} = \{P_0, P_1, ..., P_n\}$ is defined as,
\begin{align}
\label{eq:bspline}
C(t) = \sum_{i=0}^{n} N_{i,k}(t)\ P_i,
\end{align}

% They can also be seen as several Bezier curves joined end-to-end. Where a Bezier curve of degree n can only have n-1 changes of direction, a B-spline can have plenty. The B-spline defined above consists of n-k+1 Bezier curves. This removes the problem having a higher degree curve just to have several changes in direction in our curve beacuse now we can define a lower degree curve with several control points.

where, $N_{i,k}(t)$ are called the basis functions, defined recursively according to the Cox-de Boor formula \cite{cox1972numerical, de1972calculating},

\begin{equation}
\begin{aligned}
    N_{i, 0}(t) &= 
    \begin{cases}
        1 & \text{if } t_i \leq t < t_{i+1}\\
        0 & \text{otherwise}
    \end{cases}, \\
    N_{i, k}(t) &= \frac{t - t_i}{t_{i+k} - t_i} N_{i, k-1}(t) + \frac{t_{i+k+1} - t}{t_{i+k+1} - t_{i+1}} N_{i+1, k-1}(t),
\label{cox_de_boor}
\end{aligned}
\end{equation}

\vspace{0.5cm}

where, $t_i$ are called knots defined by a knot vector $\mathbf{T} = \{t_0, t_1, ..., t_s\}$, given $s=n+k+1$. 

Essentially, the knots between which a point $t$ on our curve lies, determine which basis function will affect the shape of our curve at the said point, which eventually determines where and how the CPs influence the curve (Eq. \ref{eq:bspline}). While a uniform knot vector has equally spaced knots, an open knot vector has its first and last knot values repeated $k+1$ times \cite{pieql1997nurbs}. A combination of both, known as the open uniform knot vector, is considered standard in literature, and can be easily computed from the known $k$ and $n$ values.

This definition of a B-spline curve when extended in two dimensions leads to a B-spline surface. A B-spline surface is defined as the weighted average of $(n+1)\times(m+1)$ CPs $P_{ij}$, where the weights are given by the tensor product of two 1-D basis functions $N_{i,p}(u)$ and $N_{j,q}(v)$,

\begin{align}
\label{eq:surf_def}
S(u,v) = \sum_{i=0}^{n} \sum_{j=0}^{m} N_{i,p}(u)N_{j,q}(v)\ P_{ij},
\end{align}

Similar to curves, in this case the two knot vectors $\mathbf{U}$ and $\mathbf{V}$ along the two surface parameters $u$ and $v$ can be easily pre-computed, and the surface basis functions can then be extracted using Eq. \ref{cox_de_boor}. These two surface parameters $u$ and $v$ correspond to the parametric directions of the surface.

% B-spline surfaces can be constructed by taking the tensor product of two versions of Eq. \eqref{eq:bspline}, one for each direction; a disadvantage of this method is that it requires our data to have a grid structure. \textit{Non-uniform rational B-splines} (NURBS) in particular are commonly used across fields such as computer-aided design (CAD), computer graphics and geometric modelling to approximate curves and surfaces. \textit{Rational} implies that the control points may each have a weighting which is $\neq 1$, and \textit{non-uniform} refers to the scenario in which the knots are not evenly spaced, such that more knots may be apportioned to areas of the spline which we wish to exhibit a greater degree of variability.

\begin{figure}
  \includegraphics[width=\textwidth, trim=0cm 66cm 46cm 0cm, clip]{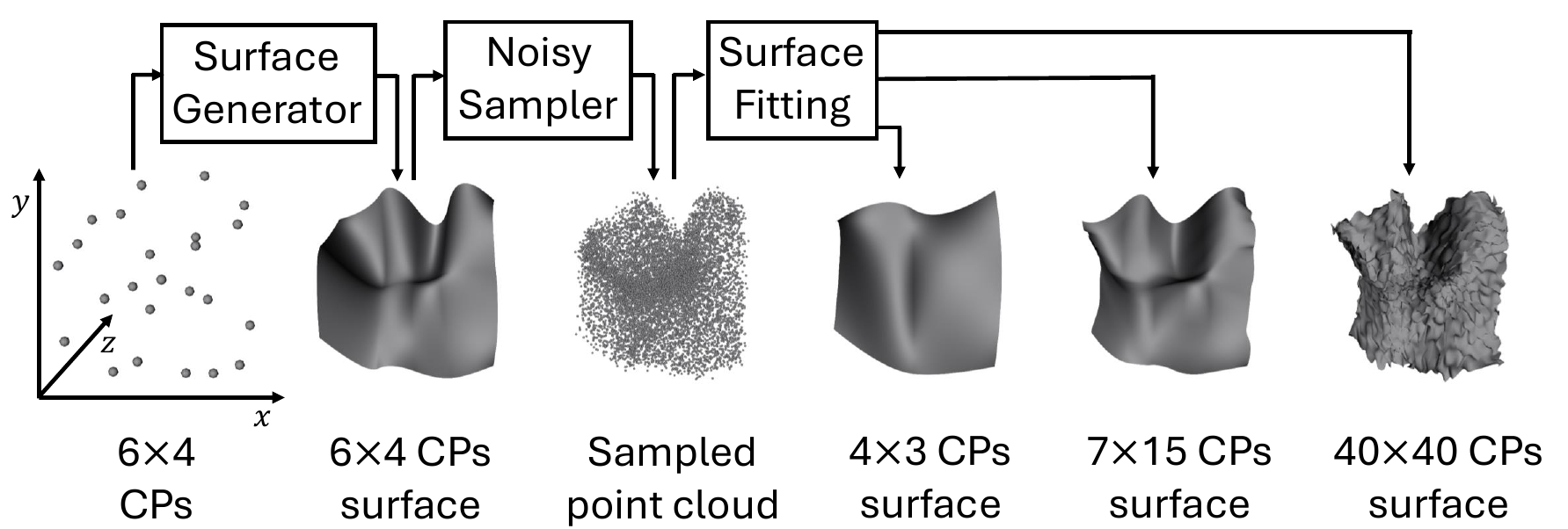}
           \caption{Our dataset creation pipeline for a single point cloud using $6\times4$ CPs spline surface is demonstrated by the first three columns. The remaining columns show the surfaces reconstructed from this point cloud using BSA \cite{pieql1997nurbs} with fixed CP grid sizes. The underfitting in $4\times3$ CPs surface and the overfitting in $7\times15$ and $40\times40$ CPs surfaces clearly explains the need of variable CP grid size estimation within a reconstruction approach.}
  \label{fig:spline}

\end{figure}

A key parameter for any B-spline-based reconstruction strategy, which considerably impacts the complexity of the generated surface from a noisy point cloud, is the number of CPs along the two parametric directions (also called the CP grid size). Typically, the values which result in an optimal trade-off between the surface smoothness and approximation quality are determined manually via intuitive trial-and-error. The importance of learning this CP grid size from the data itself is shown in Figure \ref{fig:spline}, which visualizes the consequences of the inaccurate estimation of these values. Moreover, another disadvantage of these methods lies in the fact that they require the input 3D data to have an ordered grid structure. Motivated by these challenges, our method extracts the correct CP grid size and the locations of the corresponding CPs from a given unordered non-grid structured point cloud itself.

% \subsection{Graph Convolutional Networks}

% Whilst traditional deep neural networks can be used to extract features from point clouds, they typically treat points in the same local neighbourhood independently in order to preserve permutation invariance \cite{qi2017pointnet}. This assumption is limiting as it prevents the model from capturing the underlying local geometric structure within the cloud. Motivated by this, Wang et al. introduced the \textit{dynamic graph CNN} (DGCNN), consisting of \textit{EdgeConv} layers which are capable of constructing edge features which represent the topological relationship between a given point and its nearest neighbours, whilst preserving permutation invariance \cite{wang2019dynamic}. Since this pioneering work there have been several

% % The \textit{dynamic} nature of the technique refers to the fact that rather than using a fixed graph throughout each layer of the model, the DGCNN recomputes the nearest neighbours for each point using the feature space generated by each layer of the model.

% For our work we employ a GCN (add a neural neural flowchart)

% \textcolor{red}{Maybe write about GCN a little and why we used graph based approach}

\section{Methodology}
\subsection{Dataset Creation}
\label{sec:dataset}

We generate a point cloud dataset using B-spline modeling as explained in Section \ref{sec:bg} with the implementation provided in \cite{bingol2019geomdl}. Each point cloud instance in our dataset is essentially a noisy version of a sampled B-spline surface, generated using a specific number of CPs in the two spline surface directions. To ensure non-self-intersecting spline surfaces, we first generate a uniformly-spaced $xy$-grid for CPs and then superimpose it with some Gaussian noise. However, the $z$-coordinates of the CPs vary randomly between a range, which ensures both complex individual surfaces and a comprehensive dataset. We then define the spline degrees and compute two open uniform knot vectors eventually used for B-Spline basis functions generation. The said CP grid and the basis functions generated are used to simulate a B-spline surface. This surface is first randomly sampled and then per point Gaussian noise is added. The first three figures of Figure \ref{fig:spline} visually demonstrate the generated CPs (with grid size $6\times4$), subsequent B-spline surface and the noisy sampled point cloud respectively, as an example of this data creation method. Several random samples with one CP grid size are generated. This exact procedure is repeated for several other CP grid sizes to generate a point cloud dataset with underlying B-spline surfaces of varying grid sizes. To train our model, we generate two datasets using the described method: one containing point clouds from two CP grid sizes, and another with three. By adjusting the input parameters of this generation process, users can tailor the complexity of the training datasets to suit their requirements.

% mainly two arrays: one for all the sampled noisy point clouds, and the other for all the corresponding ground truth CP grids.

% \textcolor{red}{degree 2 and training and and testing dataset specifics} 

% \begin{figure}
%   \vspace{-0.2cm}\includegraphics[scale=0.45, trim=0cm 10cm 0cm 0cm, clip]{images/dataset.pdf}\vspace{-0.2cm}
%            \caption{}\vspace{-0.2cm}
%   \label{fig:dataset}

% \end{figure}

As mentioned before, in this work we aim to predict both the number as well as the location of CPs for each test sample. This way we overcome the widely-adopted fixed CP grid size approach. The fact that the ground truth CP grid sizes are different for different instances of our dataset, makes our dataset unsuitable for a GCN model. To overcome this, we take advantage of the concept of zero-padding from machine learning-assisted 3D vision literature. Suppose, a sample from the dataset has a CP grid size of $i\times j$ then its CP grid matrix will be of size $i\times j \times 3$. We expand this matrix to a larger size of $I \times J \times 3$ such that all the new elements introduced to the original matrix are padded with zeros. Here, $I$ and $J$ are greater than the largest point cloud CP grid size in the dataset. We do this for all the samples. Additionally, to ensure better generalization after training, we randomly remove several points from each point cloud and shuffle the order of the remaining points. This enables our model to operate effectively on unordered point clouds, unlike traditional spline approximation methods that rely on ordered and gridded point sets.

\subsection{Graph-based Surface Reconstruction using B-Splines}

\begin{figure}
 \includegraphics[width=\textwidth, trim=0cm 69cm 31cm 0cm, clip]{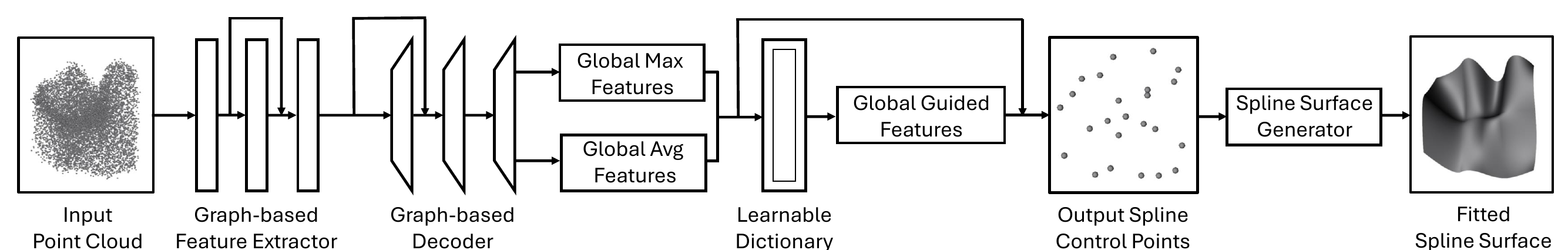} \vspace{0.05cm}
           \caption{Graph-based Surface Reconstruction using B-Splines architecture.}
  \label{fig:model}

\end{figure}

Figure \ref{fig:model} illustrates our model architecture. The pipeline integrates graph-based feature extraction, a learnable dictionary module, and a B-spline surface generator to reconstruct smooth surfaces from noisy point clouds. The process begins by feeding a noisy point cloud into several GCN layers, which extract local features. These features are aggregated into a global representation through pooling operations. This global feature vector is then refined via a dictionary-guided mechanism. The learnable dictionary captures recurring patterns between the ground truth CP values and the input cloud during training, allowing the model to learn useful priors. At inference time, these learned priors help guide accurate prediction of CPs. Finally, these CPs act as compact surface representations of the input point cloud and are used by the B-spline surface generator to reconstruct the final surface.

% \textcolor{red}{DRAFT\\
% The GCN model, see Figure \ref{fig:model}, combines deep feature extraction, skip connections, global understanding, and dictionary guiding to produce highly accurate surface reconstructions on noisy point clouds.
% The deep features are obtained by applying graph convolutions to the input point cloud data with its respective edge indexes. The input concatenates the two previous convolution output steps at the third and fifth convolutions to ensure that higher spatial information is stored.
% After the final GCN convolution is applied, a global pooling operation is used, both by average pooling and max pooling, and overall feature data is extracted from the convolutions. These global features are then passed through a dictionary-guided attention mechanism, where they are compared to learned prototypes (priors) from the training dataset, allowing the model to refine and guide its predictions using prior knowledge.}

% \subsubsection{Weighted loss function}

We employ a weighted Mean Squared Error (MSE) loss function and Adam optimizer to backpropagate the error between the predicted and the ground truth CPs. This weighted loss function gives higher importance to accurately predicting the zero-padded values in the output CPs (as discussed in Section \ref{sec:dataset}). This facilitates easy CP grid extraction later. We also incorporate skip connections between the GCN layers to preserve local structure information along the network. Our model trains on an NVIDIA GeForce RTX 4090 24GB GPU.

% \subsubsection{Step-wise Training procedure} 
% Scaling, batch size etc

\section{Experiments}
\label{sec:exp}

We conduct three sets of experiments: (1) training and testing on our own dataset with two CP grid sizes [$3\times4$, $5\times6$] containing 6000 noisy point clouds, (2) training and testing on our own dataset with three CP grid sizes [$3\times4$, $5\times6$, $3\times6$] containing 9000 noisy point clouds, and (3) evaluating the generalization abilities of our model, pre-trained on our own dataset, on random surfaces. To generate these random surfaces we use the Blender software \cite{blender} in which we manipulate gridded surfaces with geometry nodes and apply texture nodes to the normal $z$-axis of the surface. Changing the hyperparameters of these texture nodes creates complex topologies with random surface peaks and troughs. 

% Changing the hyperparameters of these texture nodes create complex topologies in the generated surfaces .
 
We compare our method against several widely-used and recently proposed surface reconstruction techniques that do not rely on normal estimation at any stage of their pipeline: (1) Ball-Pivoting (BP) \cite{bernardini2002ball}, (2) Screened Poisson (SP) \cite{kazhdan2013screened}, (3) Least Square B-spline Surface Approximation (BSA) \cite{pieql1997nurbs}, (4) Differentiable Spline Approximations (DSA) \cite{cho2021differentiable}, and (5) p-Poisson Surface Reconstruction in Curl-Free Flow (PINC) \cite{park2023p}. We preprocess the unordered point clouds in our dataset by applying gridding and ordering techniques (using Python's scientific libraries) for methods (3) and (4). This step is necessary to make the data compatible with these methods, which, as discussed in Section \ref{sec:bg}, are inherently unable to operate on unordered point sets.

To quantitatively evaluate the accuracy of our method we adopt three error metrics which basically measure the similarity between the original non-noisy point cloud and the point cloud randomly sampled from the generated surface by our method and the mentioned baselines. Lower values of these metrics indicate greater similarity between the two point clouds, reflecting more accurate surface reconstruction. We provide a brief description of these metrics along with their significance below:

Earth Mover’s Distance (EMD) is universally-adopted as not just an evaluation metric but also an optimization strategy for several point cloud processing techniques \cite{liu2020morphing, fan2017point}. It basically tries to find the best mapping function from one point cloud to another. It is recognized for delivering more reliable visual quality and offering a better intuitive measure of similarity between two point clouds when compared to the standard Chamfer Distance (CD) \cite{liu2020morphing, achlioptas2018learning, wu2021density}. 

% Its computationally expensive nature has led to faster approaches such as \cite{pele2009}

Normal Consistency (NC) measures the angular alignment between the surface normals of corresponding points in two point clouds, capturing local geometric similarity.  To compute this metric each point in one point cloud is mapped to its nearest neighbor in the other. This metric emphasizes on local smoothness and is commonly used in surface reconstruction and normal estimation tasks \cite{ben2020deepfit, li2022graphfit}.

Density-aware Chamfer Distance (DCD) is a recently proposed similarity metric \cite{wu2021density} that accounts for both the density distribution and fine structural details of the input point cloud. Its development is motivated by the limitations of the standard CD, which tends to perform poorly on datasets with outliers and noise \cite{tatarchenko2019single}. DCD is derived from the original CD, but has a higher tolerance to outliers through an approximation of Taylor expansion.

\begin{figure}
\includegraphics[width=\textwidth, trim=0cm 58.5cm 42cm 0cm, clip]{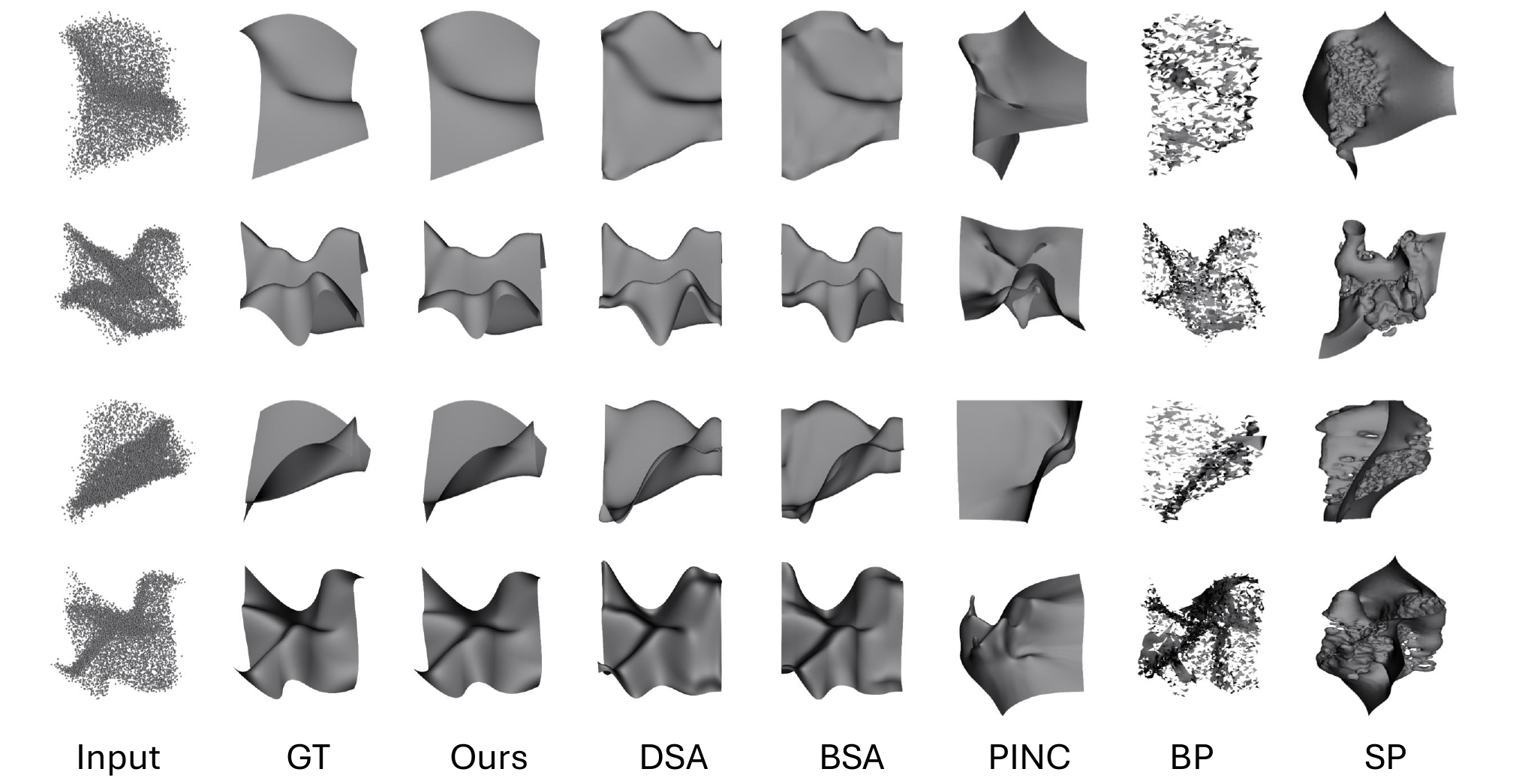}
           \caption{Surface reconstruction visualizations of four different point clouds from our dataset using our method and the other baselines. This showcases the visual superiority of our method over others.}
  \label{fig:control-point-sets}

\end{figure}

\begin{figure}

\includegraphics[width=\textwidth, trim=0cm 57cm 43cm 0cm, clip]{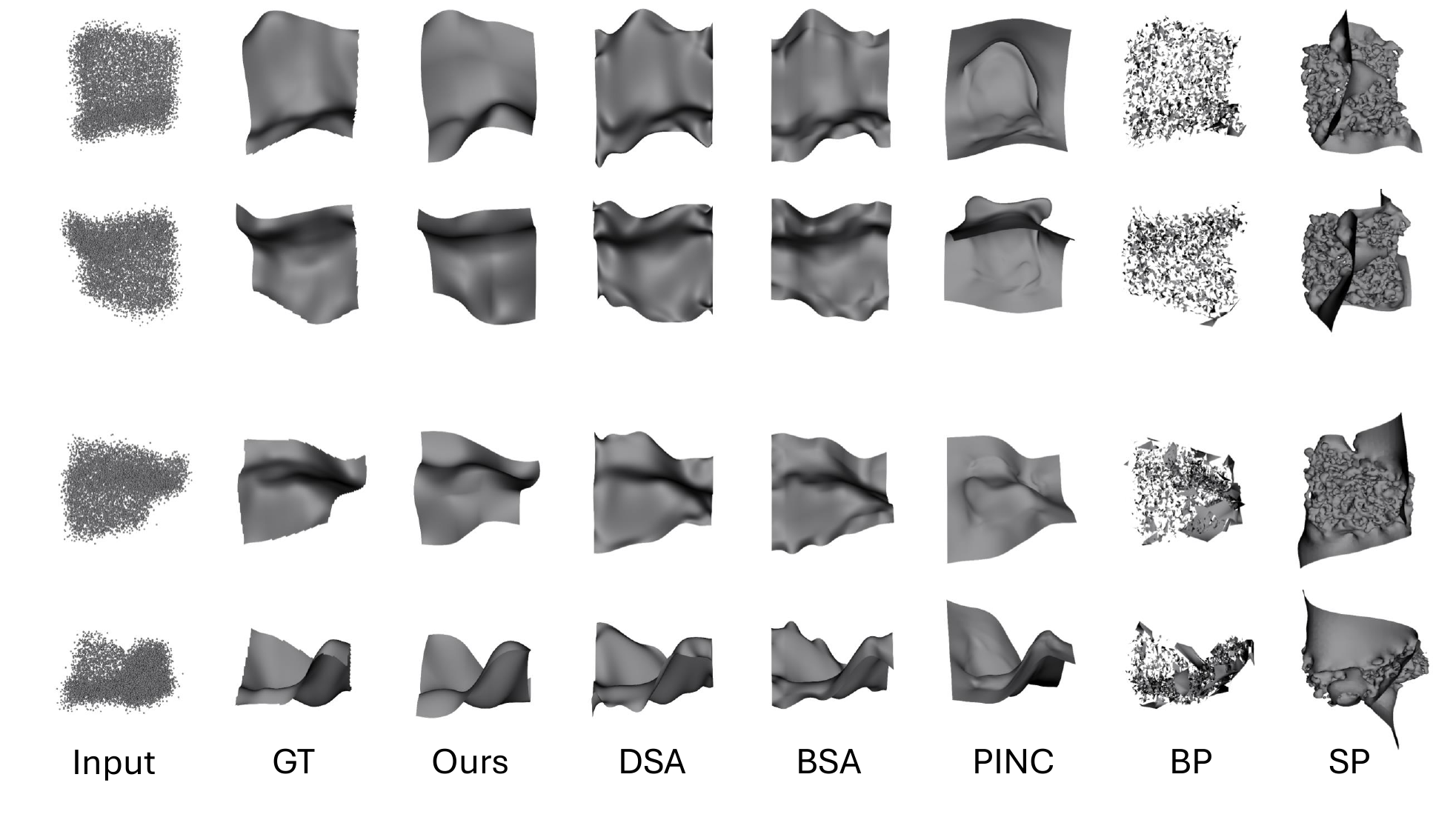} \vspace{-0.4cm}
           \caption{Surface reconstruction by our method and the other baselines experimented on two random surfaces generated using Blender \cite{blender} visualized from two different view angles. The first two rows correspond to the first surface, while the remaining two the second. This visualization highlights our method's ability to generalize to arbitrary surfaces.}
  \label{fig:blender-surfaec-test}

\end{figure}

\section{Results}

% As mentioned in the previous section, we present results for two point cloud datasets generated using method discussed in Section \ref{sec:dataset} with the potential to train on additional sizes to introduce more complexity into the model.

In this section, we present a comprehensive quantitative and qualitative evaluation of our method in comparison to the aforementioned baselines. Table \ref{tab:metrics} reports the average values of the three performance metrics NC, DCD, and EMD across two point cloud testing datasets with varying CP grid sizes. The results clearly demonstrate that our method consistently outperforms all the baselines, producing surfaces that are most similar to the ground truth (as discussed in Section \ref{sec:exp}). These quantitative results are further corroborated by qualitative visualizations shown in Figure \ref{fig:control-point-sets}, which illustrate the superior surface reconstruction quality of our approach visualized for four input point clouds. In particular, while BP is incapable of producing watertight surfaces, SP and PINC overfit the noisy point cloud to a huge extent providing less to no smoothening over the noise. Although BSA and DSA reconstruct surfaces that somewhat resemble the ground truth, their outputs also tend to overfit to the noisy input data. This behavior could be attributed to the fixed CP grid size used in their algorithms. 

Since point clouds in general vary in complexity and size, the dataset creation method described in Section \ref{sec:dataset} allows for the generation of customized training datasets tailored to specific user requirements. Based on the complexity of this training data even the model's complexity can be adjusted to ensure optimal performance.

Additionally, we experiment on some random surfaces generated by Blender (as mentioned in Section \ref{sec:exp}) as shown in Figure \ref{fig:blender-surfaec-test}. All the baselines show the same trend as observed in the previous experiments. Our model, once again, closely resembles the ground truth and gives the best visual results.

To further demonstrate the effectiveness of our approach, we include a video in the supplementary material showcasing $360^{\circ}$ visualizations of all the figures in our paper, excluding Figure \ref{fig:model}, to help readers better appreciate the accuracy and visual quality of our method as compared to other baselines.

% \begin{table*}
% \centering
% \resizebox{0.8\linewidth}{!}{  % Adjust as needed
% \begin{tabular}{c ccc ccc}
% \toprule
%  & \multicolumn{3}{c}{2 CPs grid sizes dataset} & \multicolumn{3}{c}{3 CPs grid sizes dataset} \\
% \cmidrule(lr){2-4} \cmidrule(lr){5-7}
% & DCD ($\downarrow$) & NC ($\downarrow$) & EMD ($\downarrow$) 
% & DCD ($\downarrow$) & NC ($\downarrow$) & EMD ($\downarrow$) \\
% \midrule
% BP    & 0.80946 & 0.25917 & 0.03279 & 0.80591 & 0.25967 & 0.03441 \\
% SP    & 0.77025 & 0.21204 & 0.02773 & 0.77684 & 0.20896 & 0.02904 \\
% BS    & 0.52247 & 0.10549 & 0.02577 & 0.53381 & 0.10963 & 0.02621 \\
% DSE   & 0.52422 & 0.10365 & 0.03219 & 0.52919 & 0.10410 & 0.03110 \\
% PINC  & 0.93698 & 0.32296 & 0.05924 & 0.93004 & 0.32857 & 0.05555 \\
% Ours  & \textbf{0.51779} & \textbf{0.08743} & \textbf{0.01220} & \textbf{0.46880} & \textbf{0.08084} & \textbf{0.01159} \\
% \bottomrule
% \vspace{-0.7cm}
% \end{tabular}
% }
% \vspace{0.5cm}
% \caption{Quantitative comparison of our method with different baselines experimented on 3
% 2 datasets and based on three error metrics: Earth Mover’s Distance (EMD), Normal Consistency (NC) and Density-aware Chamfer Distance (DCD). Lower are the values of these metrics higher is the surface reconstruction quality.}
% \label{tab:metrics}
% \vspace{-0.5cm}
% \end{table*}

\vspace*{0.8cm}
\begin{table*}
\centering
\resizebox{0.7\linewidth}{!}{  % Adjust as needed
\begin{tabular}{c ccc ccc}
\toprule
 & \multicolumn{3}{c}{2 CP grid sizes dataset} & \multicolumn{3}{c}{3 CP grid sizes dataset} \\
\cmidrule(lr){2-4} \cmidrule(lr){5-7}
& DCD ($\downarrow$) & NC ($\downarrow$) & EMD ($\downarrow$) 
& DCD ($\downarrow$) & NC ($\downarrow$) & EMD ($\downarrow$) \\
\midrule
BP    & 80.9 & 25.9 & 3.3 & 80.6 & 25.9 & 3.4 \\
SP    & 77.0 & 21.2 & 2.8 & 77.7 & 20.9 & 2.9 \\
BS    & 52.2 & 10.5 & 2.6 & 53.4 & 11.0 & 2.6 \\
DSE   & 52.4 & 10.4 & 3.2 & 52.9 & 10.4 & 3.1 \\
PINC  & 93.7 & 32.3 & 5.9 & 93.0 & 32.9 & 5.6 \\
Ours  & \textbf{51.8} & \textbf{8.7} & \textbf{1.2} & \textbf{46.9} & \textbf{8.1} & \textbf{1.2} \\
\bottomrule
\end{tabular}
} \vspace{0.5cm}
\caption{Quantitative comparison of our method against various baselines on two test datasets (with varying CP grid sizes) using the average values of three error metrics: Earth Mover’s Distance (EMD), Normal Consistency (NC), and Density-aware Chamfer Distance (DCD). All values are reported in units of $\times 10^{-2}$. Lower metric values indicate better surface reconstruction, with our method achieving the best performance.}
\label{tab:metrics}
\end{table*}

% \textcolor{red}{
% We present results for two sets of control point sizes, with the potential to train on additional sizes to introduce more complexity into the model. Figure \ref{fig:control-point-sets} illustrates the results of six different networks for two distinct surfaces derived from different sizes of control sets.
% When provided with a noisy point cloud, the PINC, BP, and SP methods exhibit significant failures in predicting ground truth (GT).\\
% BSA and DSA demonstrate comparable performance and output, with BSA producing steeper surface gradients.
% Our model performs closely to the desired ground truth but still exhibits deviations, particularly at the outermost points, and occasionally flattens certain peak heights.
% }

% \textcolor{red}{
% As an additional test, we utilised two randomly generated Blender surfaces. This performance test was conducted solely on our model, DSA, and BSA, based on the previous analyses. The results are presented in Figure \ref{fig:blender-surfaec-test}.\\
% DSA and BSA exhibit similar performance, producing strong curves that may be attributed to overfitting on the training dataset.\\
% Our model, once again, closely resembles the ground truth, although it exhibits deviations characterized by reduced intensities of the surface gradients and heights of peaks and troughs. These performance differences noted in Figure \ref{fig:control-point-sets} and \ref{fig:blender-surfaec-test} can also be seen in Figure \ref{fig:first}.
% }

\section{Conclusion}

Surface reconstruction from noisy point cloud data hails as one of the most challenging yet essential tasks in the computer vision community. In this paper, we try to address several drawbacks of existing reconstruction techniques such as the use of either ground truth normals or their inconsistent intermediate estimation in learning-based techniques, and fixed control points approach in case of spline-based reconstruction algorithms. We do this by leveraging a graph-based dictionary-guided learning strategy and B-spline surface fitting to obtain compact surface representations of noisy input point clouds without using any point normals. We evaluate our results based on three error metrics and compare them with several baselines. We also provide visualizations of the reconstructions obtained by our method and the other baselines used. Our method demonstrates superior performance in both the quantitative metrics and the visual quality of the reconstructed surfaces by introducing the concept of variable spline control points and eliminating the reliance on point normals.

% As a future work, our method can be 

% Our algorithm can be used patchwise for complex objects, which can be viwed as a future work
% normal estimation can be done as an added advantage (normals of b-spline) (maybe still we can do it) (we can also show results on actual point clouds like the bunny using patches of it)(add experiments done on curves too). Meshing of non-uniformly distributed point clouds leads to holes in the reconstructed surface and is a differnt problem altogether and doesnb't fall in the scope of this paper.

% \textbf{EXTRA: train and test for 4 or 5 ctrl grid sizes and report only the results of our method, do soem genralizing on blender fuinctions, do some popular functions}
\bibliographystyle{abbrv}
\bibliography{references}

\end{document}